\documentclass{article}

\pdfoutput=1
\usepackage{PRIMEarxiv}
\usepackage[utf8]{inputenc} 
\usepackage[T1]{fontenc}    
\usepackage{hyperref}       
\usepackage{url}            
\usepackage{booktabs}       
\usepackage{amsfonts}       
\usepackage{nicefrac}       
\usepackage{microtype}      
\usepackage{lipsum}
\usepackage{fancyhdr}       
\usepackage{graphicx}       
\usepackage{amsmath}
\usepackage{multirow}
\usepackage{float}

\graphicspath{{media/}}     

\title{GCBLANE: A Graph-Enhanced Convolutional BiLSTM Attention Network for Improved Transcription Factor Binding Site Prediction

}

\author{
  Jonas Chris Ferrao \\
  Department of Computer Engineering \\
  Don Bosco College of Engineering \\
  Fatorda, India \\
  \texttt{jonasferrao21@gmail.com} \\
  \And
  Dickson Dias \\
  Department of Computer Engineering \\
  Don Bosco College of Engineering \\
  Fatorda, India \\
  \texttt{diasdickson94@gmail.com} \\
  \And
  Sweta Morajkar \\
  Department of Computer Engineering \\
  Don Bosco College of Engineering \\
  Fatorda, India \\
  \texttt{sweta.morajkar@dbcegoa.ac.in} \\
  \And
  Manisha Gokuldas Fal Dessai \\
  Department of Computer Engineering \\
  Don Bosco College of Engineering \\
  Fatorda, India \\
  \texttt{manisha.faldessai@dbcegoa.ac.in} \\
}
\begin{document}
\maketitle

\begin{abstract}
Identifying transcription factor binding sites (TFBS) is important for understanding gene regulation, as these sites allow transcription factors (TFs) to bind to DNA and modulate gene expression. Despite progress in high-throughput sequencing, accurately pinpointing TFBS remains challenging due to the vast amount of genomic data and complex binding patterns. To tackle this problem, we introduce GCBLANE, a graph-enhanced convolutional bidirectional Long Short-Term Memory attention network. This model's structure combines convolutional, multi-head attention, and recurrent layers, augmented by a graph neural network, enabling the detection of crucial features for TFBS prediction. In an extensive evaluation of 690 datasets from ENCODE ChIP-seq experiments, GCBLANE exhibited exceptional performance, achieving an average AUC of 0.943. Further evaluation on 165 ENCODE ChIP-seq datasets yielded an average AUC of 0.9495, surpassing advanced models that utilise multimodal approaches, including DNA shape information. This results underscores GCBLANE's effectiveness, compared to other sophisticated methods in comprehensive assessments. By integrating graph-based learning with traditional sequence analysis techniques, GCBLANE represents a major step forward in TFBS prediction.
\end{abstract}


\section{Introduction}
Transcription Factor Binding Sites (TFBS) are specific DNA sequences that serve as anchoring points for transcription factors (TFs), which play an essential role in controlling gene expression \cite{he2020deep}. Gene expression refers to the process through which genetic information is utilised to produce functional products, such as peptides or regulatory RNA molecules. Transcription factors (TFs) are proteins that interact with DNA to influence gene activity, either promoting or inhibiting gene expression. They play a key role in facilitating the conversion of DNA into mRNA, a critical step in regulating protein synthesis within the cell. \cite{latchman1997transcription}.

Transcription factors are vital due to their ability to connect TF mutations to specific diseases and their potential as therapeutic targets \cite{semenza1998transcription}. Most transcription factors comprise a DNA-Binding Domain (DBD) that interacts with specific DNA sequences near regulatory genes \cite{zeng2020transcription}. Precise identification of Transcription Factor Binding Sites (TFBS), which span 4 to 30 base pairs in length, is crucial. Motifs are brief nucleotide sequences that are fundamental to essential biological functions, including transcription and translation. They are typically recognized by specific proteins, such as transcription factors, which bind to them and influence gene expression. These sequences are crucial for regulating genetic processes and ensuring proper cellular function \cite{rajyaguru2012scd6}. Modern sequencing technologies have generated extensive high-quality TFBS data, which is available through repositories like JASPAR \cite{rauluseviciute2024jaspar}. Yet, the considerable time and resources required for thorough TFBS identification make it challenging to address all TFBS comprehensively, leading researchers to increasingly adopt computational approaches. As a result, researchers are increasingly turning to computational methods. Current TFBS prediction techniques face challenges in accuracy and efficiency due to the vast amount of data and complex binding patterns that often fail to capture all relevant TFBS characteristics from DNA sequences. To improve prediction accuracy and efficiency, we developed a Graph-enhanced Convolutional BiLSTM Attention Network (GCBLANE). This model combines convolutional layers, recurrent layers, and the attention mechanism with a graph neural network to enhance the extraction of key features for TFBS prediction, thereby improving the model's performance on complex genomic data.

\section{Literature Review}
The rapid expansion of genomic data and improvements in processing capacity have made deep learning essential for identifying transcription factor binding sites (TFBS) \cite{zrimec2021learning}. This section examines various machine learning approaches used in TFBS prediction. DeepBind \cite{alipanahi2015predicting}, a notable deep-learning technique, employs convolutional neural networks (CNNs) and surpasses previous methods in performance. However, CNNs struggle to capture sequential dependencies. To overcome this limitation, DanQ \cite{quang2016danq} was created. This model unifies CNNs and RNNs, employing a convolutional layer to identify regulatory motifs and recurrent layers to recognise the relationships between these motifs, leading to enhanced performance compared to similar models. KEGRU \cite{shen2018recurrent}, another deep-learning approach, improves TFBS detection by integrating gated recurrent units (GRUs) with k-mer embedding. 

Compared to approaches like gkSVM \cite{ghandi2016gkmsvm} and DeepBind, KEGRU exhibits superior effectiveness. Although machine-learning models have been effective in various tasks, deep-learning techniques often surpass them in performance but struggle with interpretability. To address this challenge, attention mechanisms have been integrated into deep-learning frameworks. TBiNet \cite{park2020enhancing}, a model comprising a CNN layer, BiLSTM layer, and attention layer, utilises the attention component to concentrate on TF binding motifs, enabling it to surpass leading models such as DanQ and DeepSea. Similarly, DeepGRN \cite{chen2021deepgrn} incorporated attention mechanisms into a CNN-RNN-based structure, demonstrating competitive performance among the top methods evaluated in the ENCODE-DREAM Challenge. By utilising self-attention mechanisms and residual blocks, the Self-Attention Residual Network (SAResNet) \cite{shen2021saresnet} enhances feature extraction efficiency. Additionally, the integration of transfer learning contributes to a quicker convergence during the training process. This model delivers better accuracy compared to other methods, especially when applied to smaller datasets. By integrating both bottom-up and top-down attention modules with residual blocks and a feed-forward network, MAResNet \cite{han2022maresnet} achieved improved performance, resulting in an AUC of 0.927. MSDenseNet \cite{yin2022improving} excelled further, reaching an AUC of 0.933 through the application of fault-tolerant coding.

Recent advancements in TFBS prediction have led to the emergence of more sophisticated approaches. Notable among these are DeepSTF \cite{ding2023deepstf}, DSAC \cite{yu2023cooperation}, TBCA \cite{wang2024tbca}, and MLSNet \cite{zhang_mlsnet_2024} which employ a multimodal strategy by incorporating DNA shape features into their predictive models. While MultiTF \cite{wei2024predicting} combines graph and cross-attention networks for trimodal TFBS prediction, it also incorporates additional DNA shape and structural features. This integration of structural information with sequence data has shown promising results in improving the prediction accuracy. Another significant development is BERT-TFBS \cite{wang2024bert}, which utilises a pre-trained DNABERT-2 module. These developments emphasise the continuous need for innovations to improve model size and training efficiency. However, these approaches using language models also present computational challenges due to the high resource requirements of complex transformer-based architectures. Other existing methods rely on complex modalities, such as DNA shape, structural features, or DNA-breathing dynamics, which may increase computational overhead. While these models capture complex relationships, none have fully explored the combination of graph and sequential neural networks in a sequence-only context. Graph-based neural network approaches have shown promise in related areas of genomic research, particularly in predicting TF-target gene interactions.  Notable examples include GraphTGI \cite{du2022graphtgi} and PPRTGI \cite{ma2024pprtgi}, which leverage graph structures to capture complex relationships in genetic regulatory networks. 

To address this gap, we introduce GCBLANE, which focuses on integrating graph and sequential neural networks, specifically for sequence-based TFBS prediction, offering a simpler and more computationally efficient approach without the need for additional modalities. No existing studies have explored the combination of graph and sequential neural networks for TFBS prediction in a sequence-only context, making GCBLANE a novel and efficient approach that does not rely on additional modalities.  This unexplored avenue offers scope for developing more comprehensive and accurate models for TFBS prediction, potentially leading to breakthroughs in our understanding of gene-regulation mechanisms.

\section{Methodology}
To address the challenges in transcription factor binding site (TFBS) prediction, we propose GCBLANE, a Graph-Enhanced Convolutional BiLSTM Attention Network. This section details the key components of our methodology, including datasets, feature representation, model architecture, transfer learning, training procedure, and evaluation metrics.

\subsection{Datasets}
To evaluate the models, 690 ChIP-seq datasets and a subset of 165 ChIP-seq datasets were used. These datasets served as benchmarks for evaluating, validating and comparing predictive performance with various state-of-the-art models for TFBS prediction. The data were originally sourced from the ENCODE Dream Challenge. The 690 ChIP-seq datasets encompass approximately 25 million sequences, which spanned 91 human cell types and incorporated 161 DNA-binding proteins, including both general and sequence-specific types. Additionally, 165 ChIP-seq datasets cover 29 unique transcription factors (TFs) across 32 cell lines. All datasets were split into three portions: 70\% for model training, 10\% for model validation, and 20\% for model testing purposes.

Negative samples were created utilising the 'fasta-dinucleotide-shuffle' tool within MEME \cite{bailey2015meme}. This tool shuffles the dinucleotide sequences from the positive samples while preserving their overall nucleotide composition. The negative samples are essential for training the model to distinguish between true binding sites and random sequences.
To maintain dataset balance, the ratio of negative to positive samples was carefully controlled during data preparation. A ratio of 1:1 was maintained, where for every positive sample, a corresponding negative sample was generated. Maintaining this balance is essential for enabling the model to distinguish between positive and negative instances effectively, reducing bias, and enhancing the reliability of predictions. The positive samples are made up of DNA sequences, each 101 base pairs long, containing at least one TFBS present. The datasets can be found at \url{https://github.com/projectultra/GCBLANE}.

\subsection{Feature Representation}
Preprocessing is essential for processing DNA sequences in neural networks that require vector representations. Several encoding methods exist for this purpose. Label encoding assigns numerical values to nucleotides (e.g, A = 1, T = 2, C = 3, G = 4), but this approach may inadvertently create a hierarchy among nucleotides. An alternative method is k-mer encoding \cite{MOECKEL20242289}, which divides a sequence into k-length sub-sequences.  

\begin{figure}[h]
    \centering
    \includegraphics{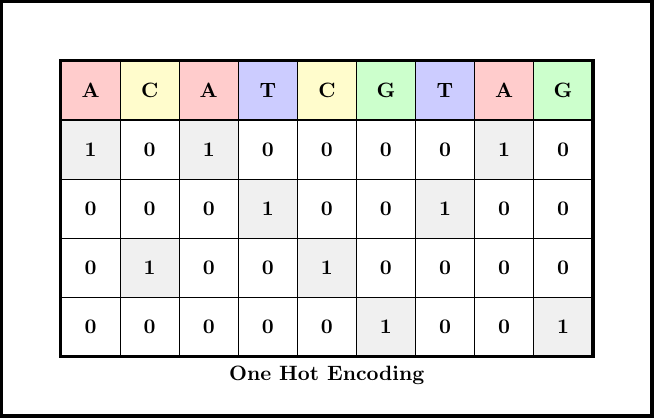}
    \caption{One-Hot Encoding}
    \label{fig:fig1}
\end{figure}

In this study, a binary vector representation known as one-hot encoding, shown in Figure \ref{fig:fig1}, was utilised to encode individual nucleotides. The encoding assigned specific binary patterns to each nucleotide: adenine (A) was represented as [1 0 0 0], thymine (T) as [0 1 0 0], cytosine (C) as [0 0 1 0], and guanine (G) as [0 0 0 1]. Using the one-hot encoding method converts each nucleotide into a vector primarily consisting of zeros, with a single bit positioned to denote the particular nucleotide being encoded. This method effectively distinguishes the four nucleotides, allowing for clear and straightforward input representation in the model. Unlike label encoding, one-hot encoding ensures that all the motifs are equally weighted. However, this approach increases the dimensionality of the input space, which can affect the computational efficiency, particularly for large datasets.

\begin{figure}[h]
    \centering
    \includegraphics{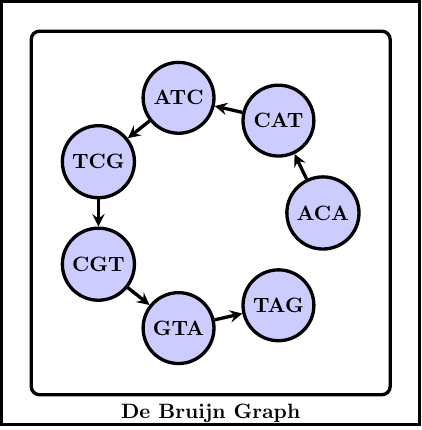}
    \caption{De Bruijn Representation of a short DNA sequence}
    \label{fig:fig2}
\end{figure}

For the graph module, a de Bruijn graph representation is employed to enhance feature extraction \cite{compeau2011apply}.
The de Bruijn graph shown in Figure \ref{fig:fig2} captures overlapping k-mers within a sequence, encoding structural relationships and dependencies in the genomic data. This allows the model to leverage additional contextual information beyond that captured by the sequential methods. By constructing a de Bruijn graph, graph neural networks (GNNs) can capture multi-hop relationships (i.e. relationships between distant k-mers) that  extend beyond immediate neighbours. In essence, GNNs provide a global context, enabling the model to look at the DNA sequence not only as a string of letters but also as a network of relationships that can influence TFBS prediction.

\subsection{Model Architecture}
\begin{figure}[H]
    \centering
    \includegraphics[height=8in]{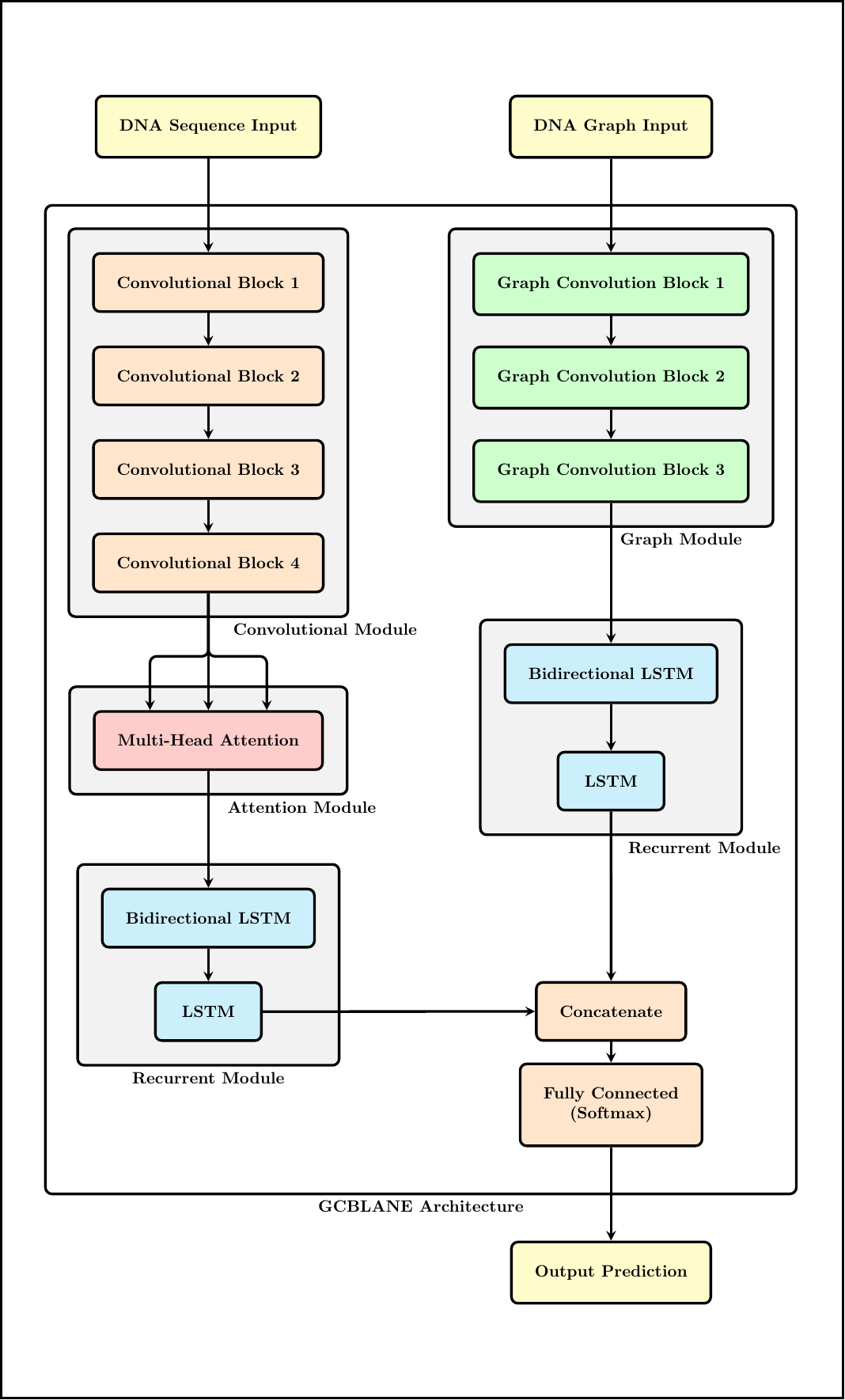}
    \caption{Model Architecture of GCBLANE}
    \label{fig:fig3}
\end{figure}

The GCBLANE Model Architecture shown in Figure \ref{fig:fig3} introduces a novel approach for transcription factor binding site prediction. To enhance the accuracy of TFBS predictions, the model incorporates a graph neural network module, which effectively represents spatial and sequential relationships in genomic data. It integrates a multi-head attention mechanism for feature extraction from sequential data, recurrent layers for capturing intermediate dependencies, and a convolutional network to address local dependencies.

\subsubsection{Convolutional Neural Network (CNN) Module:}
GCBLANE begins with a convolutional neural network (CNN) \cite{lecun1998gradient}, comprising multiple convolution layers that apply filters to extract unique features from the data. A large number of filters was selected to ensure thorough feature extraction while reducing the chance of missing key information.
The CNN layers use the PReLU activation function, an enhancement of Leaky ReLU. While Leaky ReLU allows a small gradient for negative inputs, PReLU allows the slope ($a$) of the negative part to be a learnable parameter. 
\begin{equation}
\text{PReLU}(x) = \max(0, x) + a \min(0, x)
\end{equation}
This adaptability optimises the slope during training, helping the model learn complex patterns more effectively and mitigating issues associated with fixed activation functions. To preserve critical information that might otherwise be lost, 'same' padding and a stride of one are utilised. To condense the output and lower the computational costs max pooling layers are utilised. However, the pooling layers are avoided in the initial layers to mitigate potential information loss. By focusing on the most important features from the input data, max pooling creates a more streamlined representation that enhances the overall effectiveness of the neural network. Batch normalisation is incorporated to mitigate overfitting by normalising convolutional layer outputs, adjusting each batch’s mean and variance. This stabilizes learning, accelerates convergence, and acts as a regulariser, reducing sensitivity to initial weights and promoting robust feature learning. Spatial dropout, selectively drops entire channels instead of individual neurons. This method reduces dependency on particular patterns, encouraging the model to capture broader and more adaptable representations. Together, these techniques enhance the network's ability to generalise to unseen data while maintaining a robust performance on the training set.
The operations on sequence feature of for a given CNN block is as follows:
\begin{equation}
\mathbf{P} = \text{BN} \bigg( \text{MaxPool} \Big( \text{PReLU} \big( \text{Convolution} (\mathbf{S}, \mathbf{W}_M, \mathbf{b}_M) \big) \Big) \bigg)
\end{equation}

Where $\mathbf{S}$, $\mathbf{W}_M$, and $\mathbf{b}_M$ denote the sequence features, weight matrix, and bias of the convolutional layer, respectively. $\text{PReLU}(\cdot)$ denotes the parametric ReLU activation function. $\text{MaxPool}(\cdot)$ represents the maximum pooling operation. $\text{BN}(\cdot)$ represents the Batch Normalization operation. $\text{Convolution}(\cdot)$ represents the convolution operation.

\subsubsection{Attention Network Module:}
The CNN output is passed into an attention network that uses a multi-head attention mechanism, similar to those found in transformer models \cite{vaswani2017attention}. This process derives query, key, and value representations from the CNN’s output, utilizing self-attention to process the same input across all three embeddings. The model also uses 8 attention heads. Each attention head operates separately on the input, enabling the model to capture diverse patterns and dependencies. This parallel attention strategy improves the model's capacity to extract different relationships from the data.

The attention score is computed as:

\begin{equation}
\text{Attention}(\mathbf{Q}, \mathbf{K}, \mathbf{V}) = \text{softmax}\left(\frac{\mathbf{Q} \mathbf{K}^\top}{\sqrt{d_k}}\right) \mathbf{V}
\end{equation}

where \(d_k\) is the dimensionality of the key vectors, and \(\mathbf{Q}\), \(\mathbf{K}\), and \(\mathbf{V}\) are the query, key, and value matrices, respectively.

The multi-head attention is defined as:
\begin{equation}
\text{MultiHead}(\mathbf{Q}, \mathbf{K}, \mathbf{V}) = \text{Concat}(\text{head}_1, \dots, \text{head}_h)\mathbf{W_O}
\end{equation}
where:
\begin{equation}
\text{head}_i = \text{Attention}(\mathbf{Q}\mathbf{W_i^Q}, \mathbf{K}\mathbf{W_i^K}, \mathbf{V}\mathbf{W_i^V})
\end{equation}
Where \(\mathbf{W_Q}\), \(\mathbf{W_K}\), \(\mathbf{W_V}\), and \(\mathbf{W_O}\) are learned parameters. The final output is the concatenation of outputs from all attention heads.

Furthermore, the model generates attention scores, which quantify the significance of different elements in relation to one another. These scores indicate how much focus the model places on specific features during the learning process, effectively highlighting the contributions of individual elements to the overall output. By leveraging multiple attention heads and attention scores, the model can adaptively weigh the importance of various inputs, improving its capability to extract meaningful patterns and enhance predictive accuracy.

\subsubsection{Recurrent Network Module:}
After the attention network, the data are fed into a recurrent network that starts with a bidirectional layer, allowing the model to process the input sequence in both forward and backward directions. This dual processing enables the model to capture contextual information from both ends of the sequence. Following the bidirectional layer, a Long Short-Term Memory (LSTM) \cite{hochreiter1997long} layer is employed to model long-range dependencies within the sequence data. The LSTM's architecture is particularly suited for retaining information over extended sequences, effectively addressing issues related to vanishing gradients and enabling the model to learn patterns that span across significant sequence steps. The output from the recurrent network is then channelled into a dense layer, which performs dimensionality reduction, condensing the information into a vector of size 64. This step facilitates the transition to the next stages of the model, ensuring that the essential features are retained while minimising computational complexity. 

The BiLSTM layer processes the sequence bidirectionally as follows:
\begin{itemize}
    \item \textbf{Forward Pass:} From the beginning of the sequence (\( t = 1 \)) to the end of the sequence (\( t = T \)):
    \begin{equation}
    \overrightarrow{\mathbf{c_t}} = \overrightarrow{\mathbf{f_t}} \odot \overrightarrow{\mathbf{c_{t-1}}} + \overrightarrow{\mathbf{i_t}} \odot \overrightarrow{\mathbf{\tilde{c_t}}}
    \end{equation}
    \begin{equation}
    \overrightarrow{\mathbf{h_t}} = \overrightarrow{\mathbf{o_t}} \odot \tanh(\overrightarrow{\mathbf{c_t}})
    \end{equation}

    \item \textbf{Backward Pass:} From the end of the sequence (\( t = T \)) to the beginning of the sequence (\( t = 1 \)):
    \begin{equation}
    \overleftarrow{\mathbf{c_t}} = \overleftarrow{\mathbf{f_t}} \odot \overleftarrow{\mathbf{c_{t+1}}} + \overleftarrow{\mathbf{i_t}} \odot \overleftarrow{\mathbf{\tilde{c_t}}}
    \end{equation}
    \begin{equation}
    \overleftarrow{\mathbf{h_t}} = \overleftarrow{\mathbf{o_t}} \odot \tanh(\overleftarrow{\mathbf{c_t}})
    \end{equation}

    \item \textbf{Combined Output:} The final hidden state \(\mathbf{h_t}\) at each position \( t \) is the sum of forward and backward hidden states:
    \begin{equation}
    \mathbf{h_t} = \overrightarrow{\mathbf{h_t}} + \overleftarrow{\mathbf{h_t}}
    \end{equation}
\end{itemize}

Following the BiLSTM, a unidirectional LSTM layer processes the combined output as follows:
\begin{equation}
\mathbf{c_t} = \mathbf{f_t} \odot \mathbf{c_{t-1}} + \mathbf{i_t} \odot \mathbf{\tilde{c_t}} 
\end{equation}
\begin{equation}
\mathbf{h_t} = \mathbf{o_t} \odot \tanh(\mathbf{c_t}) 
\end{equation}
where \(\mathbf{f_t}\) is the forget gate, \(\mathbf{i_t}\) is the input gate, \(\mathbf{{c_t}}\) is the candidate cell state, \(\mathbf{o_t}\) is the output gate, \(\odot\) denotes the element-wise multiplication (Hadamard product).

\subsubsection{Graph Neural Network Module:}
GCBLANE incorporates a graph neural network (GNN) \cite{scarselli2008graph} module to further enhance its capabilities. This GNN module was used to integrate additional contextual information into the model by encoding spatial information and interactions between k-mers within the genomic data. The graph module processes the graph representation to model complex relationships and dependencies that are not captured by the sequential layers alone.
Graph convolution operates on the adjacency matrix \(\mathbf{A}\) of the graph and the feature matrix \(\mathbf{X}\), where the updated feature matrix \(\mathbf{X'}\) is computed as:

\begin{equation}
\mathbf{X'} = \sigma(\hat{\mathbf{D}}^{-1/2} \hat{\mathbf{A}} \hat{\mathbf{D}}^{-1/2} \mathbf{X} \mathbf{W} + \mathbf{b} )
\end{equation}
where \(\hat{\mathbf{A}} = \mathbf{A} + \mathbf{I}\) is the adjacency matrix \(\mathbf{A}\) with added self-loops via the identity matrix \(\mathbf{I}\), \(\hat{\mathbf{D}}\) is the degree matrix of \(\hat{\mathbf{A}}\), \(\mathbf{X}\) is the input feature matrix, \(\mathbf{W}\) is the learnable weight matrix, and \(\sigma\) is the ReLU activation function.

Pooling layers reduce the graph size using MinCut clustering:
\begin{itemize}
    \item \textbf{Cluster Assignment}:
    \begin{equation}
    \mathbf{S} = \text{MLP}(\mathbf{X})
    \end{equation}
    where \(\mathbf{S}\) is computed by an MLP with softmax output.
    \item \textbf{Coarsened Features and Adjacency}:
    \begin{equation}
    \mathbf{X'} = \mathbf{S}^\top \mathbf{X}, \quad \mathbf{A'} = \mathbf{S}^\top \mathbf{A} \mathbf{S}
    \end{equation}
\end{itemize}

\subsubsection{Output Layer}
The final stage of GCBLANE consists of a dense layer with two neural units, employing a softmax activation function to generate the predicted probabilities for the presence or absence of a TFBS.

\begin{table}[H]
 \caption{GCBLANE Architecture}
  \centering
  \begin{tabular}{llll}
    \toprule
    Block & Layer & Type & Output Shape \\
    \midrule
    \multirow{6}{*}{Convolution Block 1} & Input Layer     & Input Layer     & (101, 4)     \\
                              & Convolution1D   & Convolution Layer  & (101, 256)   \\
                              & PReLU           & Activation Function & (101, 256)   \\
                              & SpatialDropout1D & Dropout Layer    & (101, 256)   \\
                              & MaxPooling1D    & Pooling Layer    & (101, 256)   \\
                              & BatchNormalization & Normalization Layer & (101, 256)   \\
    \midrule
    \multirow{5}{*}{Convolution Block 2} & Convolution1D   & Convolution Layer  & (101, 128)   \\
                              & PReLU           & Activation Function & (101, 128)   \\
                              & SpatialDropout1D & Dropout Layer    & (101, 128)   \\
                              & MaxPooling1D    & Pooling Layer    & (101, 128)   \\
                              & BatchNormalization & Normalization Layer & (101, 128)   \\
    \midrule
    \multirow{5}{*}{Convolution Block 3} & Convolution1D   & Convolution Layer  & (101, 64)    \\
                              & PReLU           & Activation Function & (101, 64)    \\
                              & SpatialDropout1D & Dropout Layer    & (101, 64)    \\
                              & MaxPooling1D    & Pooling Layer    & (50, 64)     \\
                              & BatchNormalization & Normalization Layer & (50, 64)     \\
    \midrule
    \multirow{5}{*}{Convolution Block 4} & Convolution1D   & Convolution Layer  & (50, 64)     \\
                              & PReLU           & Activation Function & (50, 64)     \\
                              & SpatialDropout1D & Dropout Layer    & (50, 64)     \\
                              & MaxPooling1D    & Pooling Layer    & (25, 64)     \\
                              & BatchNormalization & Normalization Layer & (25, 64)     \\
    \midrule
    \multirow{3}{*}{Attention Block} & Convolution1D  & Convolution Layer  & (25, 64)     \\
                              & MultiHeadAttention & Attention Layer & (25, 64)  \\
                              & Multiply        & Element-wise Multiplication & (25, 64)     \\
    \midrule
    \multirow{2}{*}{Recurrent Block 1} & Bidirectional LSTM & Recurrent Layer & (25, 64)  \\
                              & LSTM            & Recurrent Layer   & (64)         \\
    \midrule
    \multirow{2}{*}{Graph Block 1} & Graph Input Layer & Input Layer & (99, 12) \\
                              & GCNConv         & Graph Convolution Layer & (99, 128)    \\
    \midrule
    \multirow{2}{*}{Graph Block 2} & MinCutPool      & Pooling Layer    & (40, 128)    \\
                              & GCNConv         & Graph Convolution Layer & (40, 64)     \\
    \midrule
    \multirow{2}{*}{Graph Block 3} & MinCutPool      & Pooling Layer    & (12, 64)     \\
                              & GCNConv         & Graph Convolution Layer & (12, 16)     \\
    \midrule
    \multirow{2}{*}{Recurrent Block 2} & Bidirectional LSTM & Recurrent Layer & (12, 64)  \\
                              & LSTM            & Recurrent Layer   & (16)         \\
    \midrule
    \multirow{2}{*}{Output Block} & Concatenate     & Concatenation Layer & (80)         \\
                              & Dense           & Dense Layer       & (2)          \\
    \bottomrule
  \end{tabular}
  \label{tab:table}
\end{table}

\subsection{Transfer Learning}
Transfer learning has proven effective in prior state-of-the-art methods for transcription factor binding site (TFBS) prediction, enabling models to leverage generalizable features from broad datasets before adapting to specific tasks. In this work, we adopt a two-stage transfer learning strategy for GCBLANE to enhance its predictive performance across diverse datasets. This approach involves initial training on a global dataset followed by fine-tuning on each of the 690 individual ChIP-seq datasets, allowing the model to balance general pattern recognition with dataset-specific adaptation.

In the first stage, GCBLANE is pre-trained on this global dataset to capture broad, universal patterns critical for TFBS prediction, such as common motifs and sequential dependencies across TFs and cell lines. This pre-training phase leverages the model’s convolutional, recurrent, and graph neural network modules to learn a robust initial feature representation. By exposing the model to a diverse yet manageable subset of the data, it establishes a strong foundation of generalizable knowledge, reducing the risk of underfitting to the datasets during subsequent fine-tuning.

In the second stage, the pre-trained GCBLANE model is individually fine-tuned on each of the 690 specific ChIP-seq datasets. Fine-tuning adjusts the model’s weights to account for dataset-specific characteristics, such as unique TF binding preferences or cell-line-specific genomic contexts. For each dataset, the pre-trained model is initialized with the weights learned from the global dataset, and training proceeds with the dataset’s training split. This process refines the model’s ability to detect subtle, dataset-specific TFBS patterns while preserving the general features acquired during pre-training. The fine-tuning step employs a reduced learning rate (starting at 0.001 and decaying to 1 × 10\textsuperscript{-6}) to ensure stable convergence without disrupting the pre-learned representations excessively.

This transfer learning strategy offers several advantages. First, pre-training on the global dataset accelerates convergence during fine-tuning by providing a well-initialized starting point, reducing the number of epochs required for each dataset-specific model to achieve optimal performance. Second, it enhances generalization across the heterogeneous 690 datasets, as the global pre-training mitigates the impact of limited training samples in smaller datasets. Finally, by fine-tuning individually, GCBLANE adapts to the variability in TF binding behaviours and experimental conditions present in the ENCODE data, achieving superior performance compared to training from scratch on each dataset independently.

The effectiveness of this approach is evident in the results (Section 4), where GCBLANE demonstrates consistent improvements over baseline methods, particularly on small and medium-sized datasets. The transfer learning framework thus underscores GCBLANE’s scalability and adaptability, making it a practical solution for large-scale TFBS prediction tasks.

\subsection{Training Procedure}
To improve the predictive performance of the model, different hyperparameters were evaluated using a grid search during training, with a validation set used to assess their impact. All models used a batch size of 128, while the global model was trained with a learning rate of 0.001. Training spanned 10 epochs, with an early stopping mechanism applied. For each of 690 datasets the global model was fine-tuned with the learning rate starting at 0.001 and gradually decreased to as low as $1 \times 10^{-6}$ using Keras built-in callback function. The fine-tuning was conducted over 50 epochs with early stopping to avoid overfitting. The GPU runtime on Google Colaboratory, specifically a T4 GPU, was used to conduct the training process. The Adam optimizer \cite{kingma2014adam} was used, with categorical cross-entropy as the loss function.

The model was developed using TensorFlow \cite{abadi2016tensorflow} and Keras\cite{chollet2018keras}, with the graph module implemented through Spektral \cite{grattarola2021graph}. All executions were performed in a Jupyter Notebook on Google Colaboratory, which provided access to high-performance computing resources, including GPUs and TPUs, to accelerate GCBLANE's training. The full code is available on GitHub for reference and can be run in Colab for reproducibility.

\begin{table}[h!]
 \caption{Hyperparameters chosen for GCBLANE}
  \centering
  \begin{tabular}{lll}
    \toprule
    Hyper Parameter & Search Space & Selected \\
    \midrule
    Learning Rate & 0.01, 0.001, 0.0001 & 0.001 \\
    Optimizer & Adam, RMSprop, SGD & Adam \\
    Batch Size & 32, 64, 128, 256 & 128 \\
    \bottomrule
  \end{tabular}
  \label{tab:hyperparameters}
\end{table}

Table 2 summarises the hyperparameter configurations. The choice of a learning rate of 0.001 was based on preliminary experiments that indicated this rate provided a good balance between convergence speed and stability. A lower learning rate could lead to slower training without significant performance gains, while a higher learning rate risked overshooting optimal parameter values. The batch size of 128 was selected based on its impact on model performance. This size provided a good trade-off between convergence speed and the stability of training. Smaller batch sizes were tested; however, they resulted in increased training time and did not converge during training, which negatively affected the overall performance of the model. Larger batch sizes led to overfitting on the training set, which compromised the model's ability to generalise effectively.

\subsection{Evaluation Metrics}
Given that predicting transcription factor binding sites (TFBS) can be approached as a binary classification task, the model's performance was evaluated using a variety of standard classification metrics. These metrics include accuracy, precision, recall, F1 score, ROC-AUC, and PR-AUC. Each of these metrics provides valuable insights into the model's effectiveness. The terms used in these calculations are defined as follows: TP (True Positives) refers to the instances correctly identified as positive, FP (False Positives) indicates the instances incorrectly classified as positive, TN (True Negatives) denotes the instances correctly recognized as negative, and FN (False Negatives) represents the instances mistakenly classified as negative. The details of these metrics are outlined below.

    \begin{equation}
    \text{Accuracy} = \frac{\text{TP} + \text{TN}}{\text{TP} + \text{TN} + \text{FP} + \text{FN}}
    \end{equation}
    
    \begin{equation}
    \text{Precision} = \frac{\text{TP}}{\text{TP} + \text{FP}}
    \end{equation}

    \begin{equation}
    \text{Recall} = \frac{\text{TP}}{\text{TP} + \text{FN}}
    \end{equation}

    \begin{equation}
    \text{F1} = 2 \times \frac{\text{Precision} \times \text{Recall}}{\text{Precision} + \text{Recall}}
    \end{equation}

The primary comparison metric chosen was the ROC-AUC, as it assesses the ability of the model to distinguish between positive and negative classes across all possible thresholds. It is particularly useful for comparing classifiers in the context of datasets which are imbalanced, as it reflects the trade-off between the true positive rate (sensitivity) and false positive rate. ROC-AUC allows for comparison with other TFBS predictors that have used this metric, providing a standardised measure of performance. Importantly, because the ROC-AUC does not depend on a specific prediction threshold, it offers a global measure of classification performance, ensuring a robust comparison across different models.

Although ROC-AUC is widely used, PR-AUC is more informative for imbalanced datasets, where the positive class (TFBS) is relatively sparse. By concentrating on precision (how many selected items are relevant) and recall (how many relevant items are selected), PR-AUC highlights the model's ability to accurately identify positive instances. Because false positives might result in inaccurate biological interpretations, this metric is essential for TFBS prediction. PR-AUC helps to assess the model’s capability to maintain high precision across varying recall values, making it especially useful in cases where high recall is critical, such as binding site identification. Although PR-AUC has not always been reported in other papers, we include it to provide a more detailed evaluation of GCBLANE’s performance.

We evaluated our model against leading deep learning approaches, including those that exclusively use DNA sequences and others that combine DNA sequences with shape information. To ensure a fair comparison, we used identical benchmark datasets consisting of 690 and 165 ChIP-seq samples, maintaining consistent training and testing divisions.

\section{Results}
\subsection{Global Performance}
\begin{figure}[h]
    \centering
    \includegraphics[width=\linewidth]{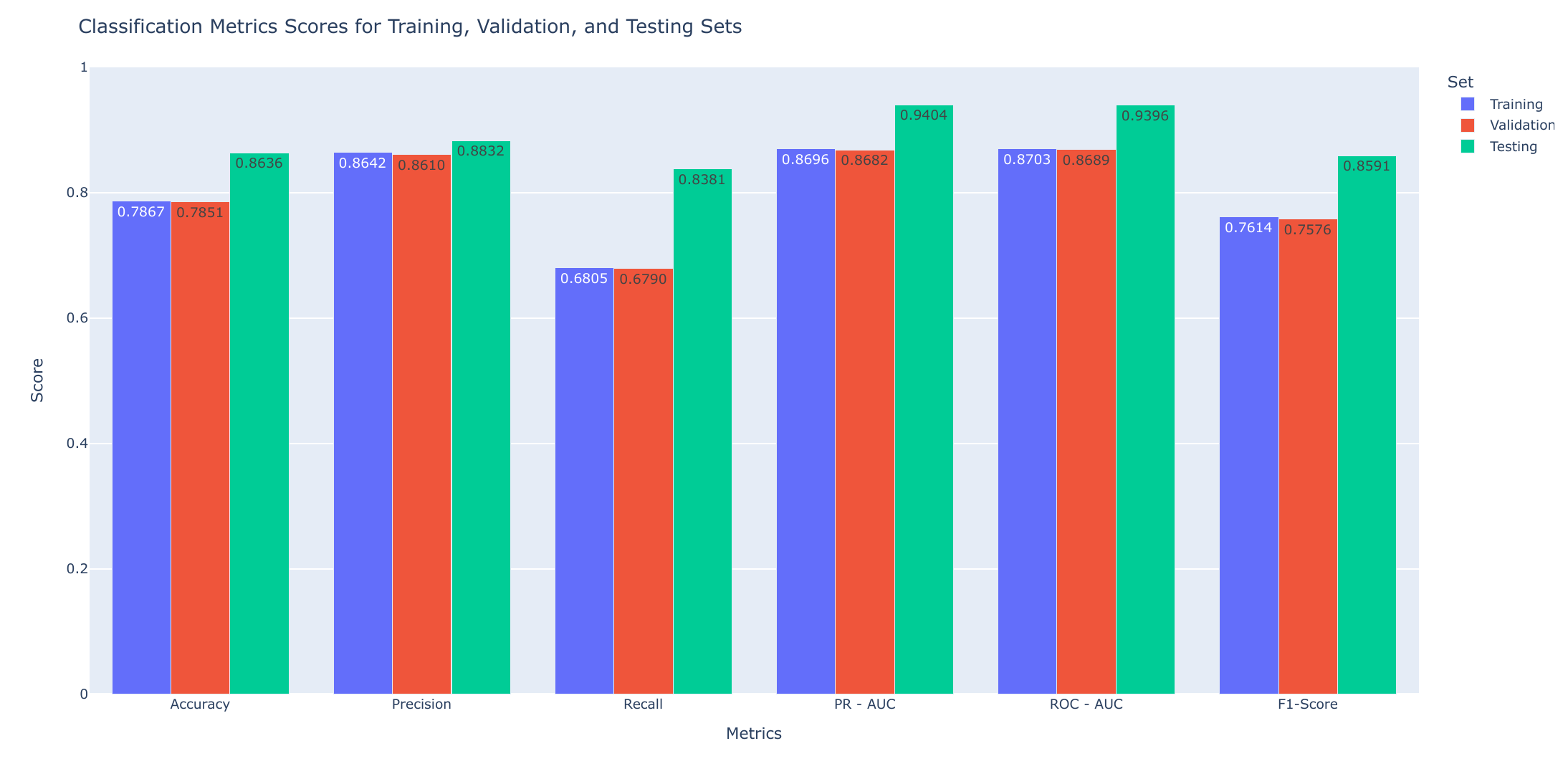}
    \caption{Visualisation of GCBLANE performance with classification metrics.}
    \label{fig:fig4}
\end{figure}

We evaluated GCBLANE's effectiveness by examining its performance on the global dataset. The results, illustrated in Figure \ref{fig:fig4}, reveal that GCBLANE attained a remarkable accuracy of 0.8636 on the test set, demonstrating its strong predictive abilities. The similar performance of the model across the validation and training sets suggests suggests effective generalization and indicates that the model avoids overfitting.

\begin{figure}[H]
    \centering
    \begin{minipage}[b]{0.45\textwidth}
        \centering
        \includegraphics[width=\textwidth]{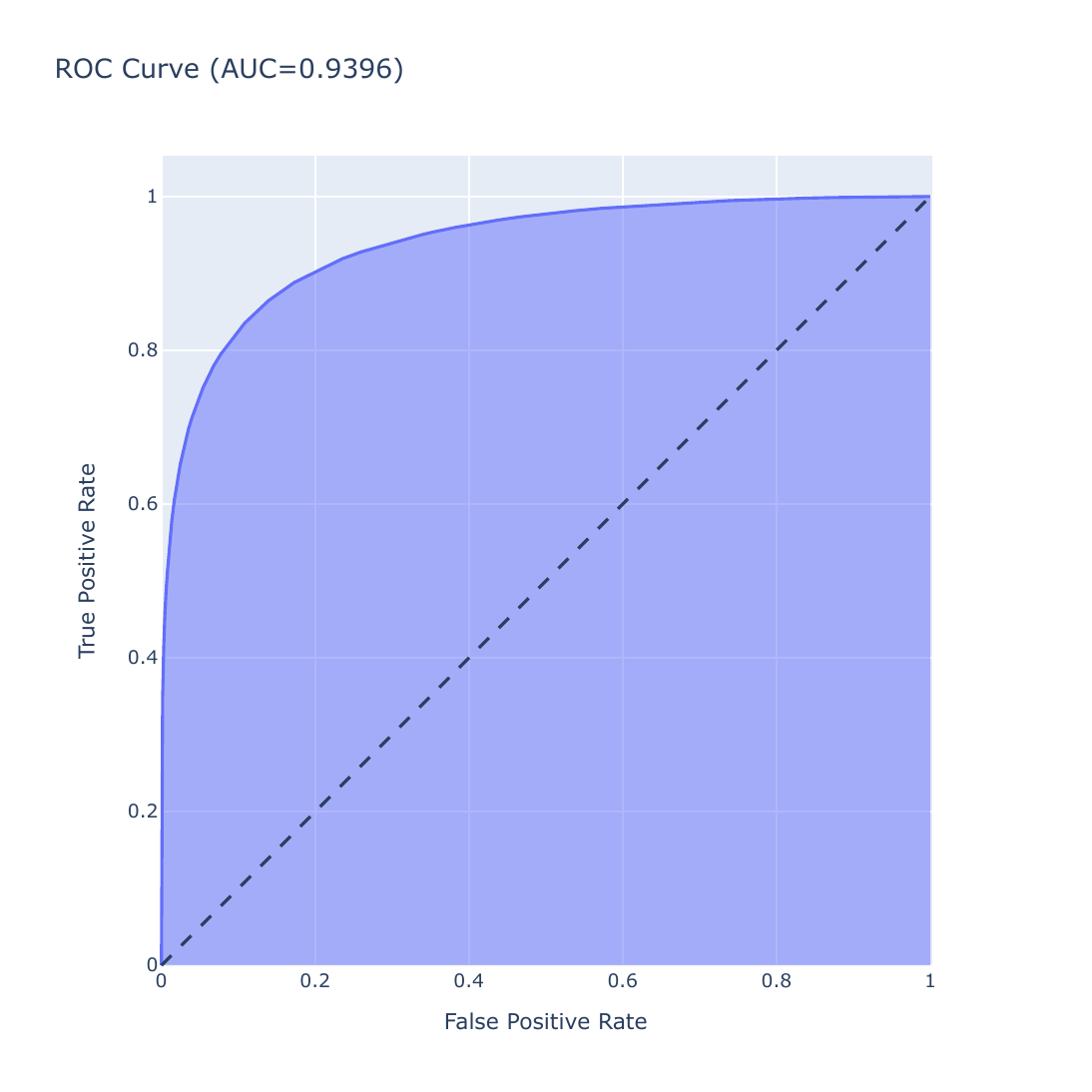}
        \caption{ROC Curve with AUC score.}
        \label{fig:fig5}
    \end{minipage}
    \hspace{0.05\textwidth} 
    \begin{minipage}[b]{0.45\textwidth}
        \centering
        \includegraphics[width=\textwidth]{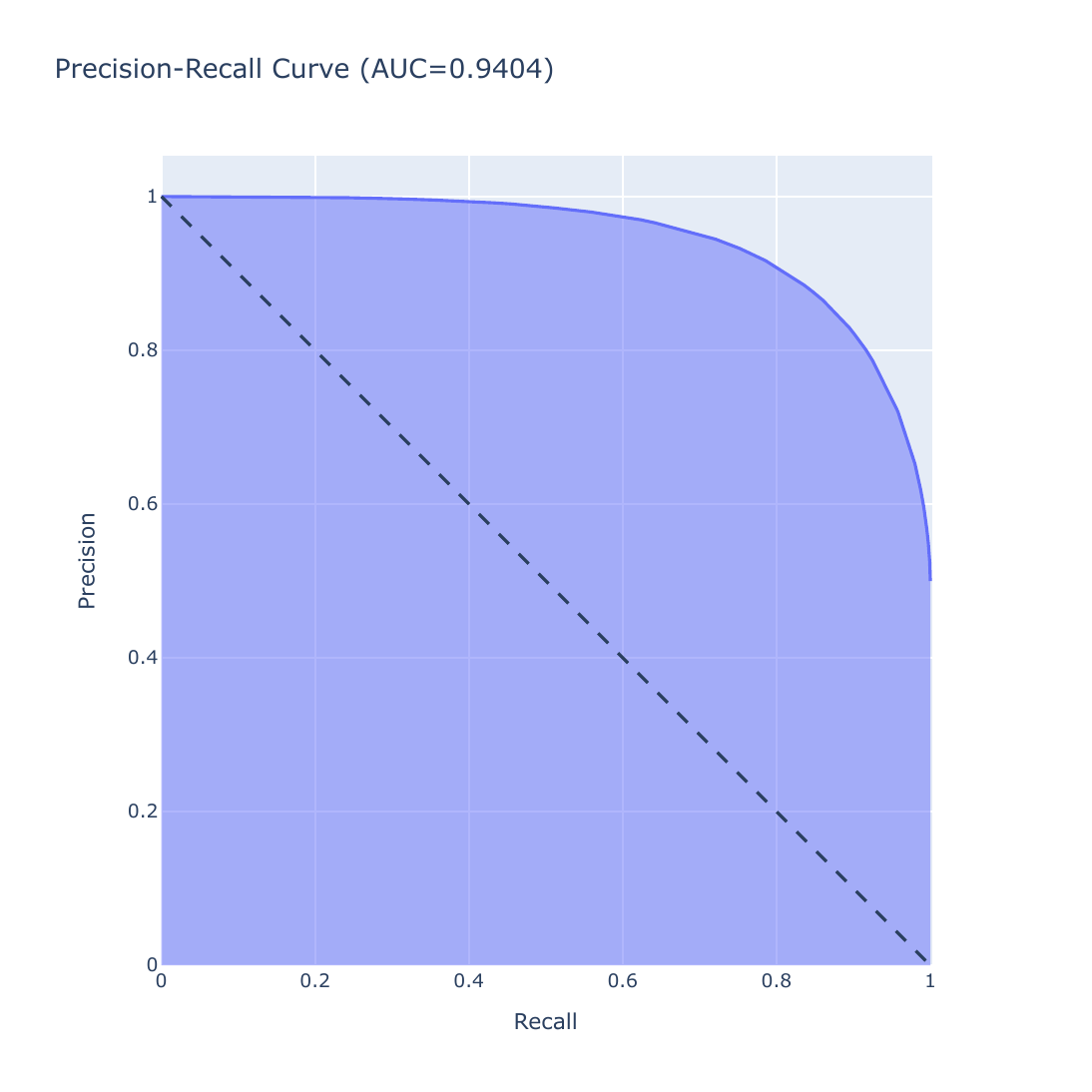}
        \caption{Precision Recall Curve with AUC score.}
        \label{fig:fig6}
    \end{minipage}
\end{figure}

To further evaluate GCBLANE’s performance, two key curves were analysed:

\textbf{Receiver Operating Characteristic (ROC) Curve:} Figure \ref{fig:fig5} shows the ROC curve, where GCBLANE achieved a high true-positive rate while maintaining a low false-positive rate. An area under the ROC curve (ROC-AUC) of 0.9396 reflects excellent discriminative ability.

\textbf{Precision-Recall (PR) Curve:} Shown in Figure \ref{fig:fig6}, GCBLANE maintained high precision across a range of recall values, indicating its ability to correctly identify positive samples without a significant drop in precision.

\begin{figure}[H]
    \centering
    \includegraphics[width=4in]{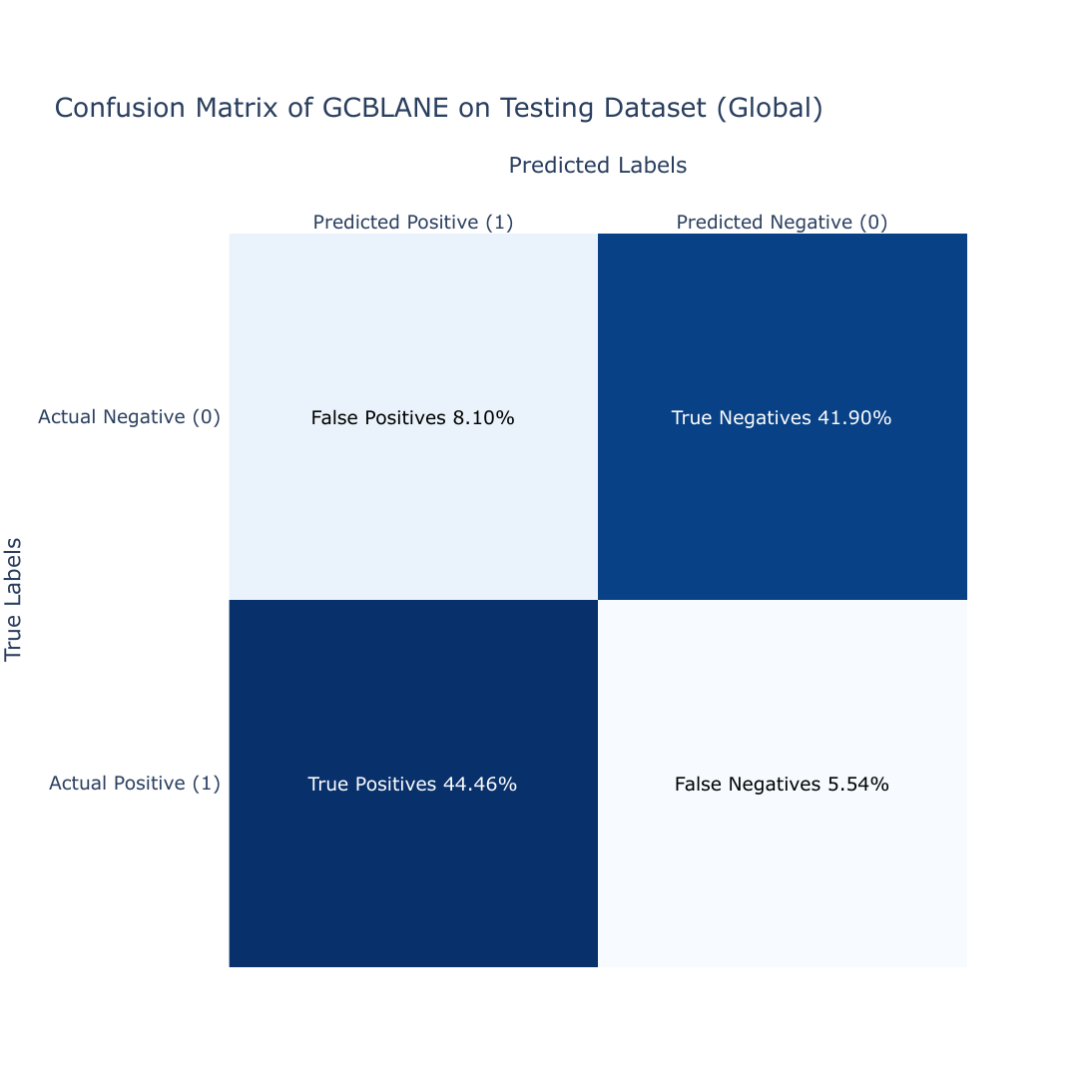}
    \caption{Confusion Matrix for GCBLANE.}
    \label{fig:fig7}
\end{figure}

Additional insights were gained from the confusion matrix (Figure \ref{fig:fig7}), which compares actual versus predicted classifications: \\
True Positives: 44.46\% of the test set, indicating strong identification of actual binding sites.\\
True Negatives: 41.90\%, showing accurate rejection of non-binding sites.\\
False Positives: 8.10\%, a relatively low rate of incorrect positive predictions.\\
False Negatives: 5.54\%, a minimal rate of missed binding sites.\\
The high true-positive rate and low false-negative rate are particularly critical for TFBS prediction, as failing to detect true binding sites can significantly affect downstream biological analyses.

\subsection{Ablation Experiments}
To understand the contribution of each component within the GCBLANE, we conducted ablation experiments by comparing it with its variants:\\
CBLANE: Excludes the graph convolutional (GCN) component from GCBLANE.\\
GNN: Uses only a standard Graph Neural Network without a recurrent, or attention network.
\begin{table}[H]
 \caption{Comparison of GCBLANE and its ablation experiments}
  \centering
  \begin{tabular}{lllllll}
    \toprule
    Model & Accuracy & ROC-AUC & PR-AUC & Precision & Recall & F1 Score \\
    \midrule
    GCBLANE & 0.8636 & 0.9396 & 0.9404 & 0.8832 & 0.8381 & 0.8591 \\
    CBLANE & 0.8143 & 0.8918 & 0.8846 & 0.8143 & 0.8143 & 0.8143 \\
    GNN & 0.6776 & 0.7476 & 0.7519 & 0.7191 & 0.5828 & 0.6418 \\
    \bottomrule
  \end{tabular}
  \label{tab:comparison}
\end{table}
The full model (GCBLANE), incorporating all components (CNN, attention, BiLSTM, and GNN), achieved the highest performance across all metrics, with an ROC-AUC of 0.9396 and PR-AUC of 0.9404. In contrast, CBLANE, a variant without the GNN module, exhibited a notable drop in performance. The ROC-AUC decreased to 0.8918 and the PR-AUC decreased to 0.8846, indicating that the GNN module plays a significant role in enhancing feature extraction, particularly in capturing structural dependencies that are not fully addressed by the sequential components alone.The GNN-only variant performed the worst, with an accuracy of 0.6776 and ROC-AUC of 0.7476. While the GNN captured some structural relationships within the data, the absence of convolutional, recurrent, and attention layers hindered the ability of the model to fully capture local motifs and sequential dependencies, which are critical for highly accurate TFBS prediction. By using de Bruijn representation, the GNN models relationships between features in a non-linear manner, capturing complex dependencies essential for biological sequences. This ability is crucial for identifying patterns that sequential models might overlook.
Analysing the contributions of each component provides insights into how GCBLANE achieves superior performance. The findings suggest that a holistic approach, which combines graph-based representation learning with traditional sequence modelling, is crucial for improving TFBS prediction accuracy.

\subsection{Overall Performance on 690 ChIP-seq Datasets}
\begin{figure}[H]
    \centering
    \includegraphics[width=\linewidth]{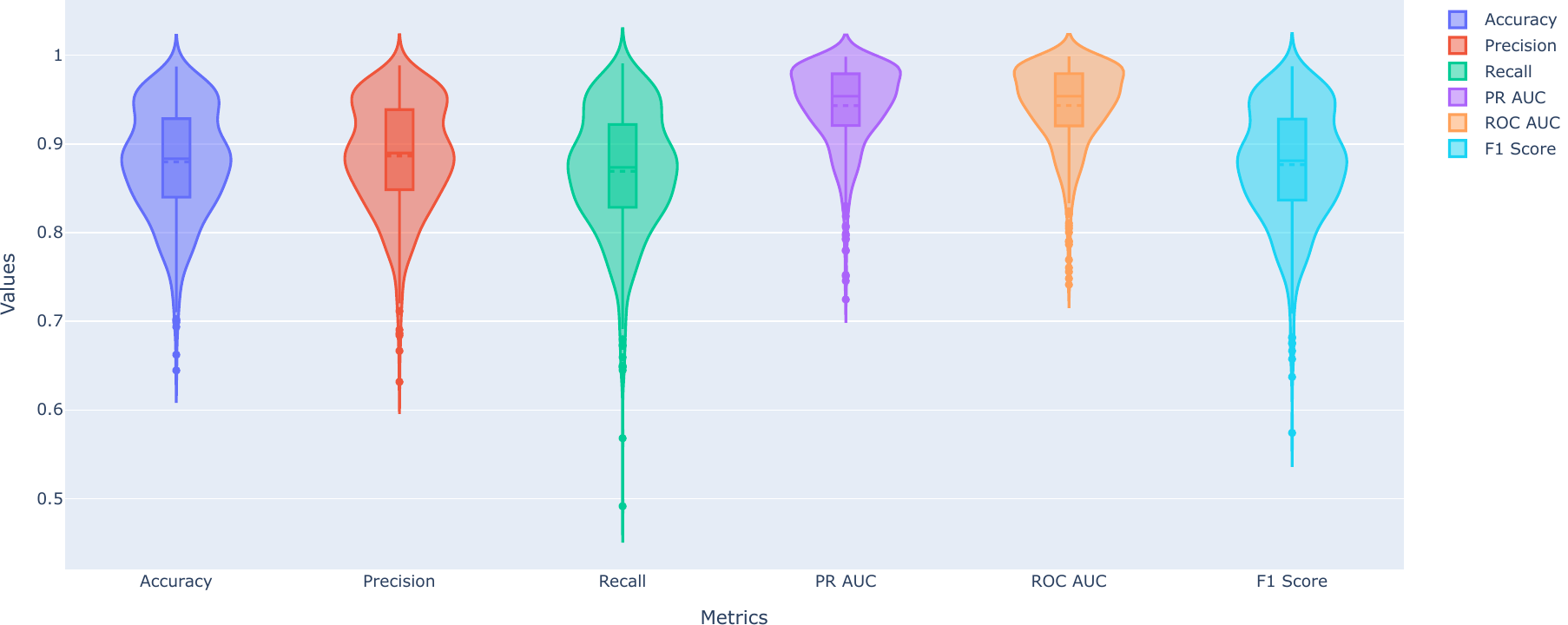}
    \caption{Violin plot illustrating GCBLANE performance on 690 ChIP-seq datasets}
    \label{fig:fig13}
\end{figure}
\begin{table}[h]
  \caption{Performance of GCBLANE on 690 ChIP-seq datasets}
  \centering
  \begin{tabular}{lcccccc}
    \toprule
    Model & Accuracy & ROC-AUC & PR-AUC & Precision & Recall & F1 Score \\
    \midrule
    GCBLANE & 0.879589 & 0.943283 & 0.943037 & 0.886301 & 0.869176 & 0.876482 \\
    \bottomrule
  \end{tabular}
  \label{tab:performance_gcblane}
\end{table}

The violin plot in Figure \ref{fig:fig13} shows the kernel density distributions for each evaluation metric, providing insights into the variability and central tendencies of model performance across the 690 ChIP-seq datasets. Each violin plot has embedded box plots indicating median and interquartile ranges. Metrics such as ROC-AUC and PR-AUC exhibit tightly concentrated distributions, reflecting consistent performance across datasets. The model demonstrates consistently high performance across all evaluation metrics on the 690 ChIP-seq datasets, as shown in Figure 8 and Table 4. The model achieves an accuracy of 0.8796 with ROC-AUC and PR-AUC scores of 0.9433 and 0.9430, respectively, indicating excellent classification capability.

\subsection{Comparison with other predictors}
We evaluated GCBLANE's effectiveness by comparing its results with those of other cutting-edge techniques across datasets of varying sizes as shown in Table 4, Table 5 and Table 6. The categorization of the 690 ChIP-seq datasets adheres to the same classification criteria employed by all other state-of-the-art methods:
\begin{itemize}
    \item Small Datasets: Fewer than 3000 training examples
    \item Medium Datasets: Between 3000 and 30,000 training examples
    \item Large Datasets: Greater than 30,000 training examples
\end{itemize}

\begin{table}[h]
  \caption{Performance comparison across different datasets}
  \centering
  \begin{tabular}{lcccc}
    \toprule
    Method & All datasets & Small & Medium & Large \\
    \midrule
    GCBLANE & 0.943 & 0.904237 & 0.93031 & 0.9732 \\
    MSDenseNet & 0.933 & 0.897 & 0.921 & 0.973 \\
    MAResNet & 0.927 & 0.883 & 0.914 & 0.972 \\
    SAResNet & 0.920 & 0.876 & 0.907 & 0.966 \\
    HOCNN & 0.887 & 0.821 & 0.868 & 0.957 \\
    Expectation- Lou & 0.881 & 0.835 & 0.859 & 0.947 \\
    CNN Zeng & 0.875 & 0.818 & 0.850 & 0.953 \\
    \bottomrule
  \end{tabular}
  \label{tab:performance_comparison}
\end{table}

GCBLANE exhibited superior performance, particularly on small and medium datasets, where it achieved the highest accuracy across all methods, indicating its robust predictive capabilities in scenarios with limited training data. On every dataset, we can observe that GCBLANE performs better than any other model, proving its exceptional performance.

Next, using the 165 datasets, we compare GCBLANE with other cutting-edge techniques that make use of multimodal techniques like DNA Shape.

\begin{figure}[H]
    \centering
    \includegraphics[width=\linewidth]{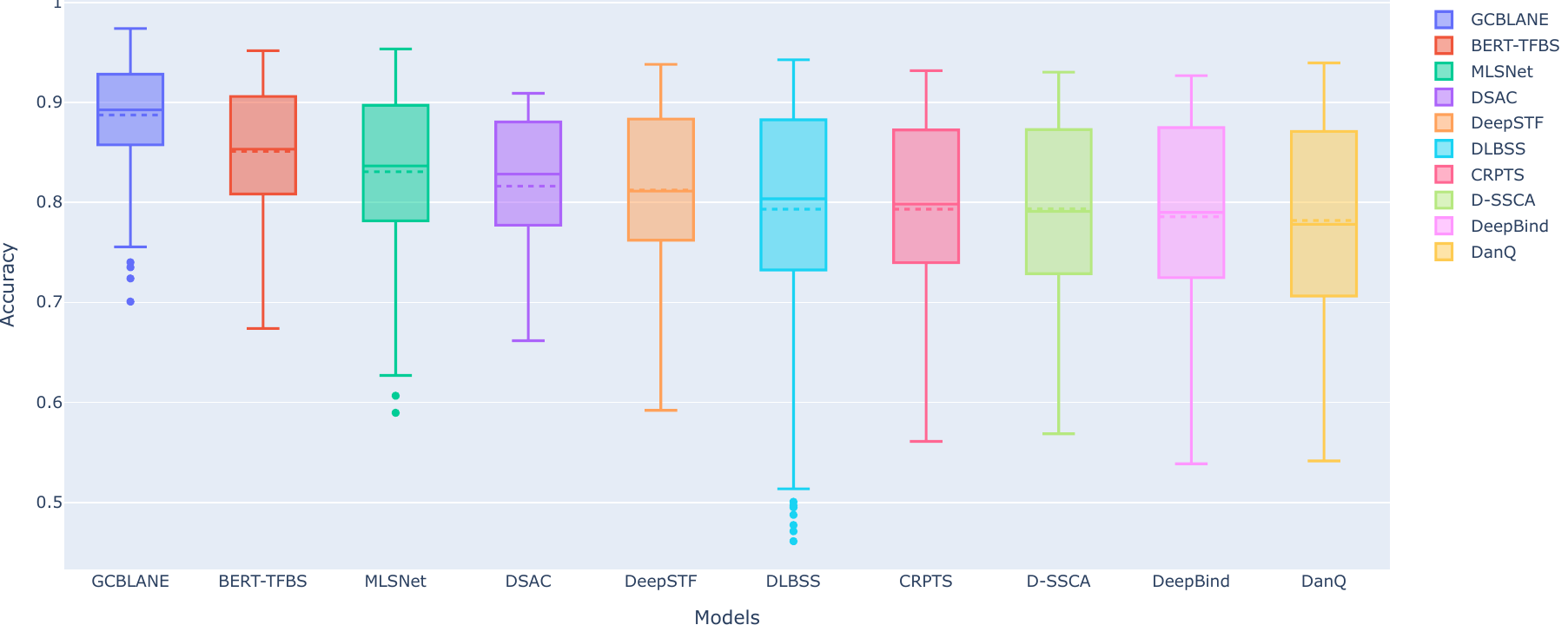}
    \caption{Distribution of accuracy scores on 165 ChIP-seq datasets for state-of-the-art models.}
    \label{fig:fig8}
\end{figure}
\begin{figure}[H]
    \centering
    \includegraphics[width=\linewidth]{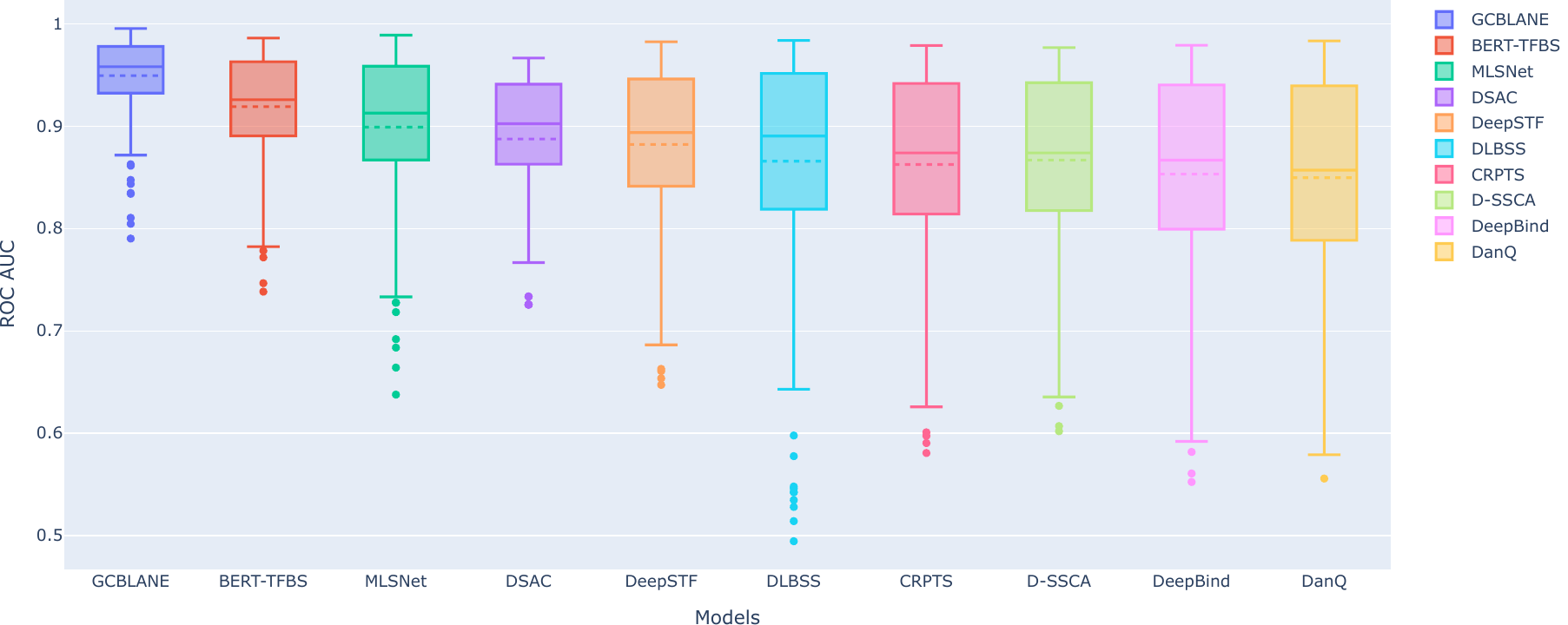}
    \caption{Distribution of PR-AUC scores on 165 ChIP-seq datasets for state-of-the-art models.}
    \label{fig:fig9}
\end{figure}
\begin{figure}[H]
    \centering
    \includegraphics[width=\linewidth]{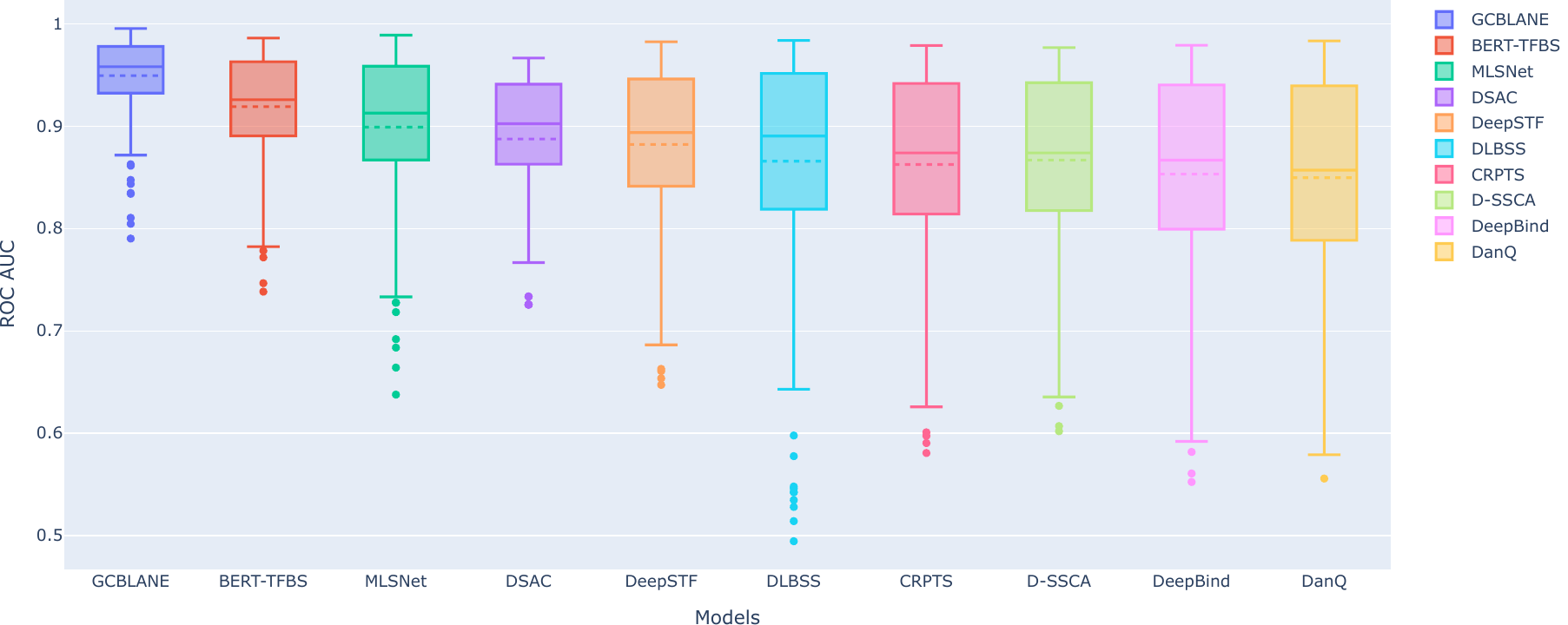}
    \caption{Distribution of ROC-AUC scores on 165 ChIP-seq datasets for state-of-the-art models.}
    \label{fig:fig10}
\end{figure}
The distribution of accuracy, ROC-AUC, and PR-AUC—three classification metrics—for GCBLANE and alternative methods is depicted in Figure \ref{fig:fig8}, \ref{fig:fig9} and \ref{fig:fig10} based on the testing sets of 165 ChIP-seq datasets. Across all three metrics, GCBLANE outperforms previous approaches, showing higher values in terms of its mean, median, quartiles, maximum, and minimum. This demonstrates GCBLANE's exceptional performance compared to existing methods.

Sequence-based models like BERT-TFBS, DLBSS, DanQ, and DeepBind lag behind GCBLANE, with lower accuracy and AUC values, reflecting their limited ability to capture relevant features compared to GCBLANE's approach. For instance, BERT-TFBS achieves an accuracy of 0.851 and a PR-AUC of 0.920, which is significantly below GCBLANE's performance.

\begin{table}[H]
  \caption{Performance of different models on DNA Information}
  \centering
  \begin{tabular}{lcccc}
    \toprule
    Model & Accuracy & ROC-AUC & PR-AUC & DNA Information Type \\
    \midrule
    GCBLANE & 0.887 & 0.9495 & 0.949 & Sequence \\
    BERT-TFBS & 0.851 & 0.919 & 0.920 & Sequence \\
    MLSNet & 0.831 & 0.899 & 0.904 & Sequence, Shape \\
    TBCA & 0.823 & 0.894 & 0.899 & Sequence, Shape \\
    DSAC & 0.816 & 0.887 & 0.891 & Sequence, Shape \\
    DeepSTF & 0.814 & 0.883 & 0.890 & Sequence, Shape \\
    D-SSCA & 0.793 & 0.867 & 0.871 & Sequence, Shape \\
    CRPTS & 0.793 & 0.862 & 0.867 & Sequence, Shape \\
    DLBSS & 0.793 & 0.865 & 0.871 & Sequence \\
    DanQ & 0.782 & 0.849 & 0.855 & Sequence \\
    DeepBind & 0.785 & 0.853 & 0.858 & Sequence \\
    \bottomrule
  \end{tabular}
  \label{tab:dna_models}
\end{table}

Even sequence-shape-based models, such as MLSNet, TBCA, and DSAC, also fall short of GCBLANE's performance despite incorporating additional DNA shape information. MLSNet, for example, reaches an accuracy of 0.831, which is much lower than GCBLANE. These results suggest that the incorporation of shape information alone does not compensate for the enhanced feature extraction mechanisms utilized by GCBLANE.

\subsection{Cell Lines and Transcription Factors Results}

\begin{figure}[H]
    \centering
    \includegraphics[width=\linewidth]{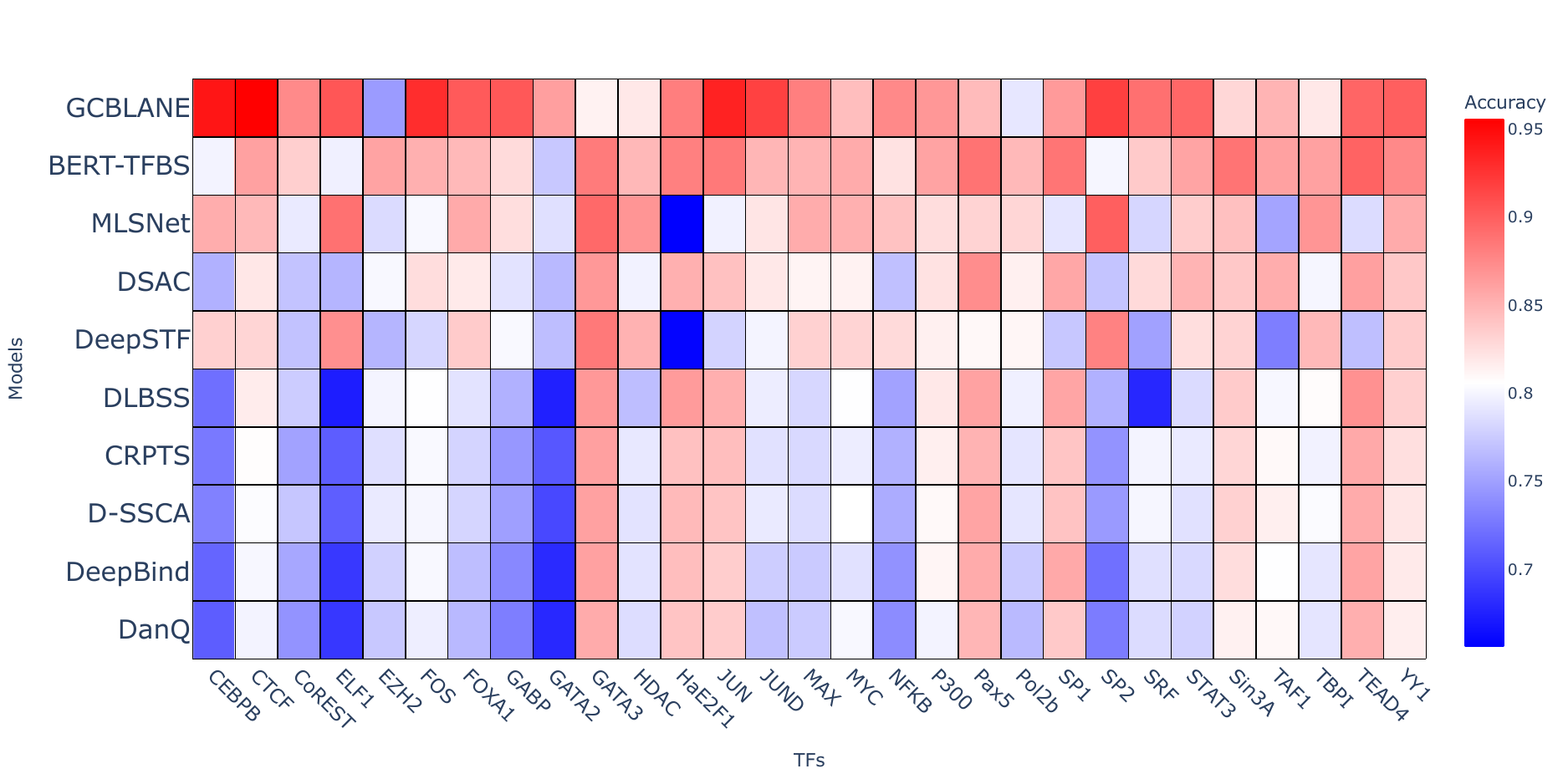}
    \caption{Heatmap of Model Accuracy Across different Transcription Factors.}
    \label{fig:fig11}
\end{figure}

\begin{figure}[H]
    \centering
    \includegraphics[width=\linewidth]{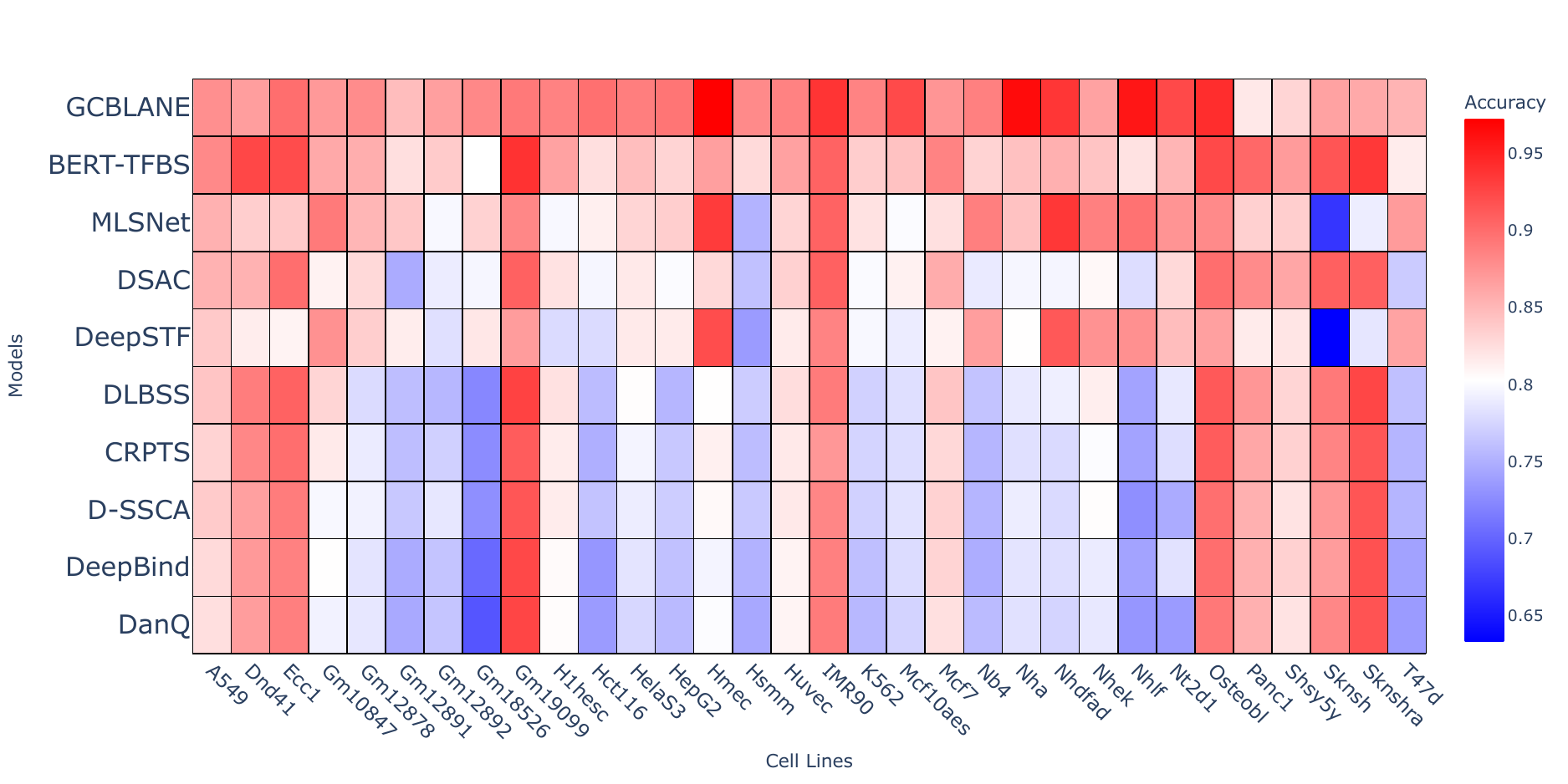}
    \caption{Heat map of Model Accuracy Across different Cell Lines.}
    \label{fig:fig12}
\end{figure}

To comprehensively evaluate the performance of GCBLANE across various cell lines and transcription factors (TFs), we analysed the results of the 165 ChIP-seq datasets, partitioned by 32 cell lines and 29 TFs. Heat maps (Figures \ref{fig:fig11} and \ref{fig:fig12}) illustrate the accuracy of GCBLANE compared to other models on these datasets, with a more intense red colour representing higher accuracy scores.

GCBLANE exhibits superior performance across most cell lines and TFs, further underscoring its robust generalization capabilities. Specifically, GCBLANE outperforms other models on key TFs, including CEBPB, CTCF, FOS, and JUN, achieving consistently high accuracy. Similarly, the model excels on cell lines such as Nha and Nhlf, where previous methodologies have struggled to deliver comparable results.

Additionally, a consistent trend is observed across all models when applied to different cell lines and TFs. Certain TFs and cell lines yield high accuracy across most models, while others exhibit uniformly lower performance. This trend highlights the influence of factors such as dataset size, the biological characteristics of specific TFs, and the unique properties of cell lines on transcription factor binding site prediction. GCBLANE’s ability to address these challenges effectively demonstrates its advantage over alternative approaches.

To summarise, the experimental outcomes showcase the exceptional effectiveness of the newly introduced GCBLANE when compared to all current benchmark models across the 165 ChIP-seq datasets. The research indicates that highly accurate predictions for TFBSs can be obtained without the need for multimodal approaches or the use of computationally demanding language models.

\section{Conclusion}
In this study, we created and applied GCBLANE, a cutting-edge deep learning model for DNA sequence Transcription Factor Binding Site (TFBS) prediction. GCBLANE integrates convolutional, multi-head attention, recurrent layers, and a graph neural network module, thereby significantly enhancing its predictive capabilities. The inclusion of the GNN module addressed the limitations of traditional sequential models by enabling the representation of complex spatial relationships in DNA sequences. This novel combination of sequential and graph-based learning provides GCBLANE with a unique advantage over the existing methods.  When compared to previous state-of-the-art techniques, the model exhibits notable gains in computing efficiency and performance on small and medium-sized datasets. On 690 ChIP-seq datasets, GCBLANE performed better than any existing technique for TFBS prediction. Its exceptional performance over other top models, including those that use DNA shape characteristics, was further validated by experiments conducted on 165 ChIP-seq datasets. Even though GCBLANE performs well, there is room for development, particularly when dealing with huge datasets, where it hasn't been able to outperform alternative techniques. One possible enhancement could be the incorporation of DNA shape profiles and structural information into the model.

\section*{Acknowledgments}
This research did not receive any specific grant from funding agencies in the public, commercial, or not-for-profit sectors.
\bibliographystyle{unsrt}  
\bibliography{references}

\end{document}